\documentclass[letterpaper, 10 pt, conference]{ieeeconf}  

\IEEEoverridecommandlockouts                              

\usepackage{amsmath} 
\usepackage{amssymb}  
\usepackage{bm}
\usepackage{multirow}
\usepackage{makecell}
\usepackage{bbding}

\usepackage{graphics} 
\usepackage{epsfig} 
\let\labelindent\relax

\usepackage{mathtools}
\usepackage{enumitem}
\usepackage{times} 

\usepackage[utf8]{inputenc} 
\usepackage[T1]{fontenc}    
\usepackage{hyperref}       
            
\usepackage{url}            
\usepackage{booktabs}       
\usepackage{amsfonts}       
\usepackage{nicefrac}       
\usepackage{microtype}      
\usepackage{xcolor}         
\usepackage{cleveref}
\usepackage{hyperref}
\usepackage{graphicx}
\usepackage{float}
\usepackage{multirow}
\usepackage{diagbox}

\usepackage{algorithm}
\usepackage{algpseudocode}
\usepackage{arydshln}
\usepackage{marvosym}

\usepackage{multirow}
\usepackage{listings}
\usepackage{graphicx}
\usepackage{siunitx} 
            
\definecolor{dkgreen}{rgb}{0,0.6,0}
\definecolor{gray}{rgb}{0.5,0.5,0.5}
\definecolor{mauve}{rgb}{0.58,0,0.82}
\newcommand{\rebuttal}[1]{\textcolor{black}{#1}}
\title{\LARGE \bf
RGBGrasp: Image-based Object Grasping by Capturing Multiple Views during Robot Arm Movement with Neural Radiance Fields
}

\author{Chang Liu*, Kejian Shi*, Kaichen Zhou*, Haoxiao Wang, Jiyao Zhang, Hao Dong
\thanks{Chang Liu, Jiyao Zhang and Hao Dong are with CFCS, School of CS, Peking University and National Key Laboratory for Multimedia Information Processing.
Kejian Shi is with Imperial College London.
Kaichen Zhou is with Oxford University.
Haoxiao Wang is with Tianjin University of Technology.
Jiyao Zhang is also with Beijing Academy of Artificial Intelligence (BAAI).}
\thanks{* indicates equal contribution}
\thanks{
Corresponding to hao.dong@pku.edu.cn}
}

\begin{document}
\maketitle


\begin{abstract}





Robotic research encounters a significant hurdle when it comes to the intricate task of grasping objects that come in various shapes, materials, and textures. 
Unlike many prior investigations that heavily leaned on specialized point-cloud cameras or abundant RGB visual data to gather 3D insights for object-grasping missions, this paper introduces a pioneering approach called RGBGrasp. 
This method depends on a limited set of RGB views to perceive the 3D surroundings containing transparent and specular objects and achieve accurate grasping.
Our method utilizes pre-trained depth prediction models to establish geometry constraints, enabling precise 3D structure estimation, even under limited view conditions. Finally, we integrate hash encoding and a proposal sampler strategy to significantly accelerate the 3D reconstruction process.
These innovations significantly enhance the adaptability and effectiveness of our algorithm in real-world scenarios.
Through comprehensive experimental validations, we demonstrate that RGBGrasp achieves remarkable success across a wide spectrum of object-grasping scenarios, establishing it as a promising solution for real-world robotic manipulation tasks. The demonstrations of our method can be found on: \url{https://sites.google.com/view/rgbgrasp}
\end{abstract}



\section{Introduction}


Robotic object grasping is a pivotal challenge in robotics due to the diverse morphology and material properties of real-world objects. 
The task necessitates accurate 3D geometry estimation, particularly for subsequent grasping pose determination. 
Although point-cloud sensors are prevalent in current approaches \cite{ten2017grasp, liang2019pointnetgpd}, they encounter difficulties with transparent and specular objects, leading to unreliable 3D environmental information capture in these scenarios. Additionally, point clouds derived from these sensors often suffered from low-resolution 3D observations and issues like noises\cite{camuffo2022recent}.

 \begin{figure}[t]
\begin{center}
    \includegraphics[width=\linewidth]{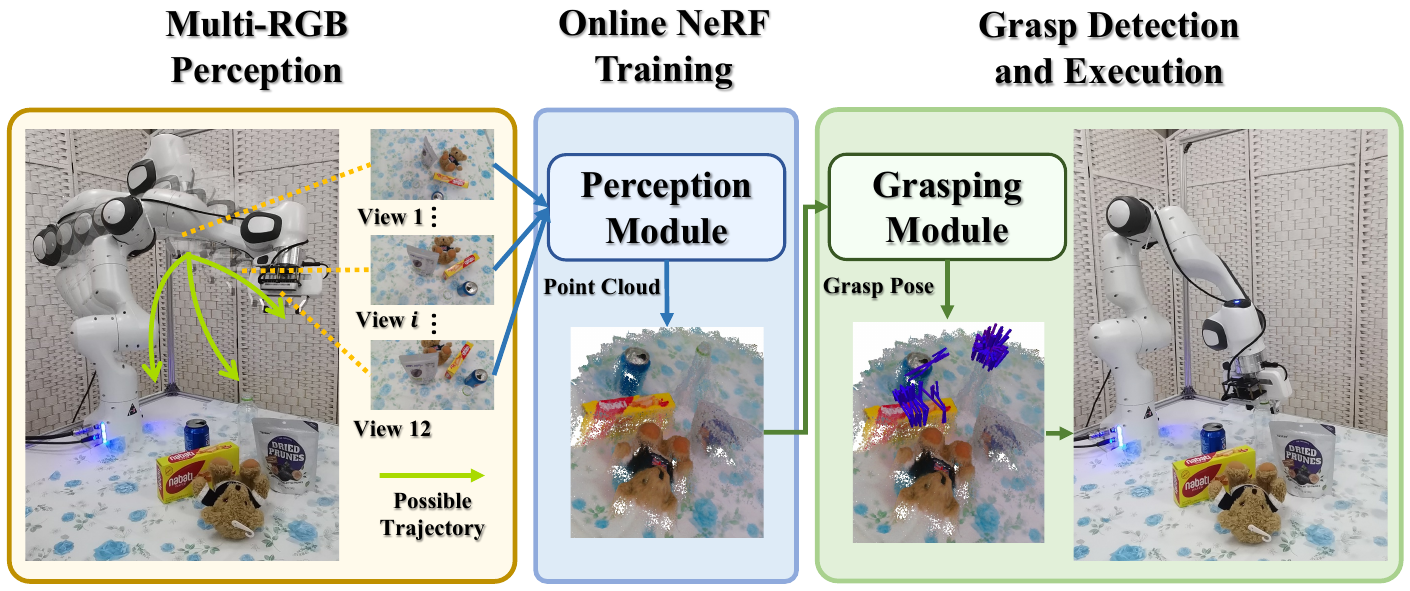}
\end{center}
\vspace{-0.3cm}

\caption{\textbf{Overview of RGBGrasp.} We introduce a novel approach capable of reconstructing the 3D geometric information of a target scene using views acquired during standard grasping procedures. Our method is not limited to fixed viewpoints and can flexibly work with partial observations in different trajectories based on the environmental requirements.}
\label{fig:teasor}
\vspace{-0.5cm}
\end{figure}

In light of the aforementioned considerations, there has been a growing interest in leveraging the RGB modality to enhance the process of object grasping. One avenue of research focuses on achieving 3D scene perception from a single image. For instance, \cite{zhu2014single} employs a single RGB image for the estimation of 3D object detection and pose, while \cite{zhai2023monograspnet} utilizes a single image to predict 2D keypoints along with their corresponding depth information.

Besides the straightforward rationale underpinning single-image-based grasping models, the task of inferring 3D geometric information from a single image poses inherent challenges, notably due to the absence of scale information \cite{zhou2022devnet}. Consequently, contemporary approaches predominantly harness multi-view RGB data to reconstruct comprehensive 3D scenes. For instance, \cite{dai2023graspnerf} leverages the neural radiance field (NeRF)~\cite{nerf2020} framework to aggregate information from multiple viewpoints, thereby achieving precise object grasping for items exhibiting diverse material properties.


Despite the impressive achievements of contemporary multi-view-based grasping algorithms, they encounter persistent challenges. Firstly, accurately reconstructing explicit geometric representations for 3D scenes that incorporate objects with varying material properties, such as transparent and specular objects, remains a difficult task. Secondly, these algorithms typically necessitate an unnatural multi-view data collection process, exemplified by \cite{dai2023graspnerf} requiring a series of frames captured by orbiting around the target scene. For example when objects are positioned on shelves or against walls, where capturing images from certain angles is impractical.
Thirdly, multi-view-based methods typically entail extended learning durations, limiting their applicability in real-world scenarios.

Given the challenges faced by prior algorithms, we introduce RGBGrasp, an innovative multi-view grasping algorithm aimed at reconstructing high-resolution 3D target scenarios.
In our configuration, an eye-on-hand camera captures multiple views of the scene as the gripper approaches the object during manipulation. 
Although these views exhibit minor disparities, our objective is to attain both precise scene reconstruction and efficient reconstruction speed.
Precisely, we integrate prior geometric information from pre-trained monocular depth estimation networks with the 3D reconstruction capabilities of neural radiance fields to achieve precise 3D scene reconstruction in scenarios with limited views. 
Additionally, we utilize hash encoding and a novel sampling proposal network to enhance the expediency of the learning process for targeted 3D scenarios.
In summary, our contributions can be encapsulated as follows:
\begin{itemize}

    \item We introduce depth rank loss~\cite{wang2023sparsenerf}, aided by a trained depth estimation model, enabling precise geometric estimation under sparse view scenarios. \rebuttal{This approach effectively overcomes the limitations of conventional NeRF methods, particularly in environments with constrained viewing angles.}
    \item We integrate a hash-encoding strategy with a proposal sampler strategy, resulting in accelerated 3D scenario reconstruction.
    \item We conducted experiments for both multi-view and sparse-view scenarios, encompassing grasping and perception tasks. RGBGrasp consistently exhibited superior performance when compared to recent approaches.
\end{itemize}


\section{Related Work}   
\subsection{Robot Grasping}  
Robot grasping plays a crucial role in enabling robots to interact with objects in the physical world. Extensive research has been dedicated to enhancing the grasping capabilities of robots over the years. GraspNet~\cite{fang2020graspnet} and AnyGrasp~\cite{fang2023anygrasp} are two notable examples of grasp detection networks trained to propose viable grasp poses based on the local depth geometry observed from RGB-D cameras. Additionally, GSNet~\cite{wang2021graspness} introduced a novel geometrical metric called graspness, which improves the speed and accuracy of grasp detection by a large margin.



\subsection{Neural Radiance Field}
Neural radiance field (NeRF)~\cite{nerf2020} is an advanced neural network-based approach for scene representation and synthesis. It generates photorealistic views by modeling scenes as continuous 3D functions from a set of input images and corresponding camera parameters. NeRF's ability to handle complex geometry and lighting conditions makes it valuable for applications in virtual reality~\cite{deng2022fov, kondo2022deep}, robotics~\cite{blukis2022neural, yan2023efficient, simeonov2022neural}, image synthesis~\cite{schwarz2020graf, gu2021stylenerf}, and more. 

\subsection{NeRF in Robotics}
 Recent studies have demonstrated the efficacy of NeRFs as a promising scene representation in several domains within robotics research. Specifically, NeRF has shown potential in robot navigation~\cite{sucar2021imap, adamkiewicz2022vision}, manipulation~\cite{li20223d, yen2022nerf}, object detection~\cite{hu2023nerf}, multi-object dynamics modeling\cite{2022-driess-compNerf}, and 6D pose tracking~\cite{chidananda2022pixtrack}. DexNeRF~\cite{ichnowski2021dex} proposes to construct the NeRF representation from 49 RGB images taken from different viewpoints around the scene and use volumetric rendering to produce a depth map from a top-down view for grasping. GraspNeRF~\cite{dai2023graspnerf} leverages generalizable NeRF to enable the model to directly work on novel scenes without retraining and downgrading the number of input views needed for NeRF to 6 sparse views. However, these works still require taking RGB images 360 degrees around the scene to construct the global context. These methods will fail in some scenarios when the visual horizon is limited or viewing from the back of the scene is impractical (\emph{e.g.}, grabbing objects from a shelf). We propose RGBGrasp and build on advantages in NeRF to solve this problem when only partial observations are available.

\section{Method}

\subsection{Problem Statement and Method Overview}

\begin{figure*}[t]
\begin{center}
    \includegraphics[width=0.7\linewidth, height=6cm]{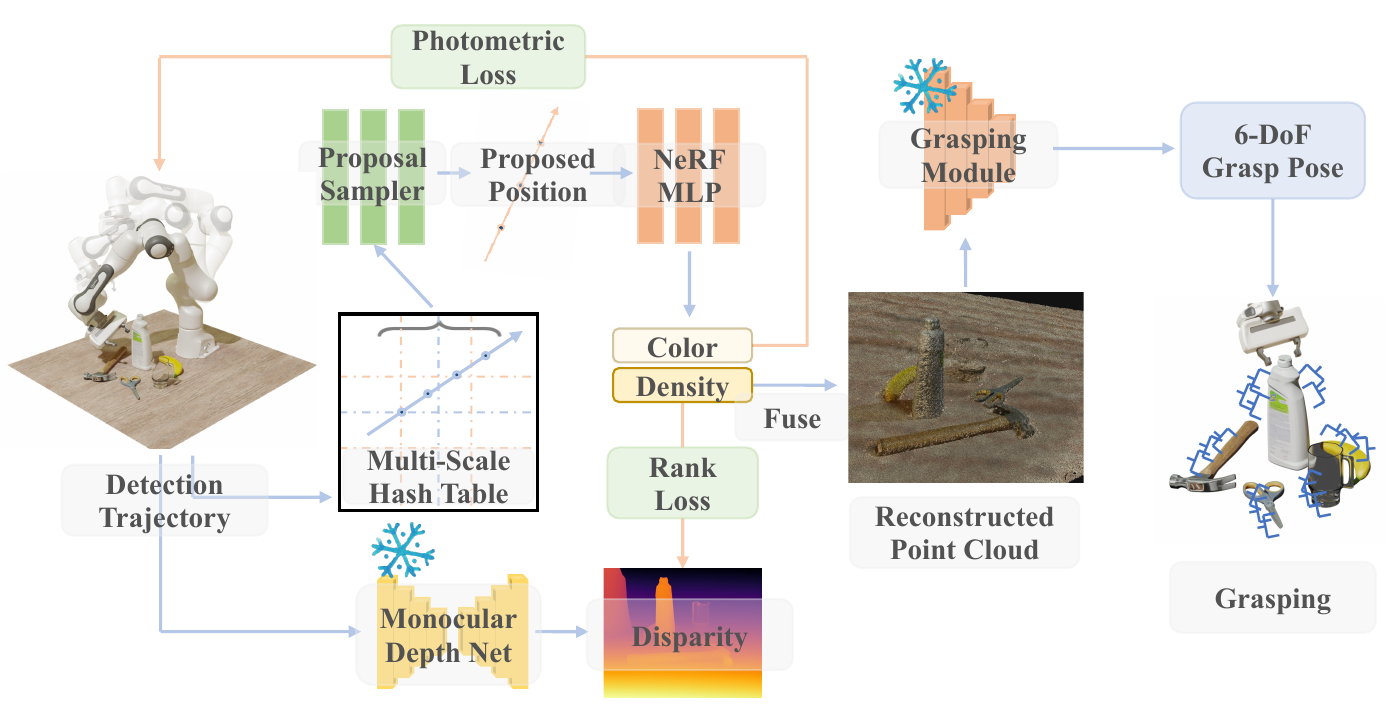}
\end{center}
\vspace{-0.7cm}
\caption{\textbf{
RGBGrasp Pipeline.
} 
The robot employs an approaching trajectory to the objects when capturing multiple views to build a multi-scale hash table. Subsequently, a proposal sampler is trained to enhance the precision of sampling positions for a subsequent fine predictor. This predictor provides color and density data for individual points, where the density information contributes to the construction of the final point cloud. This resultant point cloud serves as input for a pre-trained grasping module to predict a 6-DoF grasp pose. Throughout the optimization procedure, we maintain a fixed state for the monocular depth network and the grasping module, rendering them non-trainable components. In contrast, the Hash Table, Proposal Sampler, and NeRF MLP are actively updated and subject to the learning process.}
\vspace{-0.6cm}
\label{fig:method}
\end{figure*}

As shown in Fig.~\ref{fig:method}, RGBGrasp employs an eye-on-hand configuration by strategically mounting one RGB camera on the wrist of the robot gripper. The camera continuously captures RGB images from various angles as the robot arm directly approaches the target objects during the grasping process, those RGB images are used to progressively refine depth information. Similar to Evo-NeRF~\cite{kerr2022evo}, we assume that the objects and the robot are placed on the same planar workspace, and the objects keep rest with stable poses within the reach of the robot arm. Additionally, camera extrinsic at each step can be calculated by gripper-to-camera transform.

The robot continuously takes observation of the scene and learns about the geometry while its arm gets closer proximity to the objects. Then it predicts some grasp poses and directly executes the grasp action based on the grasp pose of the highest confidence.

Specifically, at each time step $t$, given the current and previous recorded RGB images $\mathcal{I}=\{I_0,I_1,...,I_t\}$ with their corresponding camera extrinsics $\mathcal{P}=\{P_0,P_1,...,P_t\}$, we incrementally train a NeRF-Based model from scratch to estimate the depth of the scene. Subsequently, we employ state-of-the-art grasp pose detection methods, like GraspNet~\cite{Fang2020CVPR}, to leverage the reconstructed depth data for predicting optimal grasp poses.


\vspace{-0.15cm}
\subsection{NeRF Preliminaries}
NeRF~\cite{nerf2020} learns a 5D neural radiance field which maps the spatial coordinate $(x,y,z)$ and view direction $(\theta, \phi)$ to the volume density $\sigma$ and color $\boldsymbol{c}$. The expected color $C(\boldsymbol{r})$ of camera ray $\boldsymbol{r}(t)=\boldsymbol{o}+t\boldsymbol{d}$ with near and far bounds $t_n$ and $t_f$ is:
\vspace{-0.2cm}
\begin{equation}\label{eq:volumerendering}
\begin{multlined}
C(\boldsymbol{r})=\int_{t_n}^{t_f}T(t)\sigma(\boldsymbol{r}(t))\boldsymbol{c}(\boldsymbol{r}(t),\boldsymbol{d})\rm{d}t\\
    \text{where}\ T(t)=\exp\left(-\int_{t_n}^t\sigma(\boldsymbol{r}(s))\rm{d}s\right)
\end{multlined}
\end{equation}

To discretize the computation process, NeRF approximates the expected color $\hat{C}(\boldsymbol{r})$ as:
\vspace{-0.2cm}

\begin{equation}\label{eq:volumerendering-discrete}
\hat{C}(\boldsymbol{r})=\sum_{i=1}^{N}T_i(1-\exp(-\sigma_i\delta_i))\boldsymbol{c}_i
\end{equation}

where $T_i=\exp(-\sum_{j=1}^{i-1}\sigma_j\delta_j)$ and $\delta_i=t_{i+1}-t_{i}$. Here $\sigma_i$ represents the transparency of the point, and NeRF leverages alpha-compositing with alpha values $\alpha_i=1-\exp(-\sigma_i\delta_i)$.

Similar to the color information computation, the depth information could be computed as:

\vspace{-0.15cm}

\begin{equation}\label{eq:depthrendering-discrete}
    {Z}(\boldsymbol{r})=\sum_{i=1}^{N}\alpha_i\prod_{j=1}^{i-1}(1-\alpha_j)\delta_i
\end{equation}

\subsection{Scene Depth Reconstruction}
Since the robot arm performs a top-down grasping task and directly takes a trajectory from the higher initial position to the scene, the gripper does not need to pre-rotate around the scene for a full or half circle in order to get observation of the whole scene. Therefore, the pictures that the on-hand RGB camera captures have a much smaller angular distribution range in the horizontal plane, which means the model may not have sufficient information about the geometry of the scene. 
Furthermore, the continual expansion of the training dataset can introduce instability into the training process and adversely affect the accuracy of the final reconstructed scenarios. Additionally, prevailing NeRF-based grasping algorithms are often characterized by extended training times, impeding their practical applicability.

\subsubsection{Geometry Regularization} 
To address inaccurate depth estimation resulting from sparse views and diverse material properties of objects, RGBGrasp integrates monocular depth information $\hat{Z}$ into the Neural Radiance Field (NeRF) learning process. However, utilizing the depth information obtained from a pre-trained monocular model directly in the training process is challenging due to the absence of scale information. To mitigate this issue, we employ the rank loss function to leverage the predicted depth information.

Despite the absence of scale in $\hat{Z}$, it offers precise relative depth relationships between different points. Drawing inspiration from SparseNeRF \cite{wang2023sparsenerf}, RGBGrasp incorporates a rank loss to address this aspect, expressed as follows:
\begin{equation}
    \mathcal{L}_R = \sum_{\hat{Z}_{k2} \leq \hat{Z}_{k1}} \max( Z_{k2} - Z_{k1} + \epsilon, 0),
\end{equation}
where $k1$ and $k2$ are indices of 2D pixel coordinates, and $\epsilon$ is a small margin that allows limited depth ranking errors.

\begin{figure*}[t]
\begin{center}
    \includegraphics[width=0.9\linewidth]{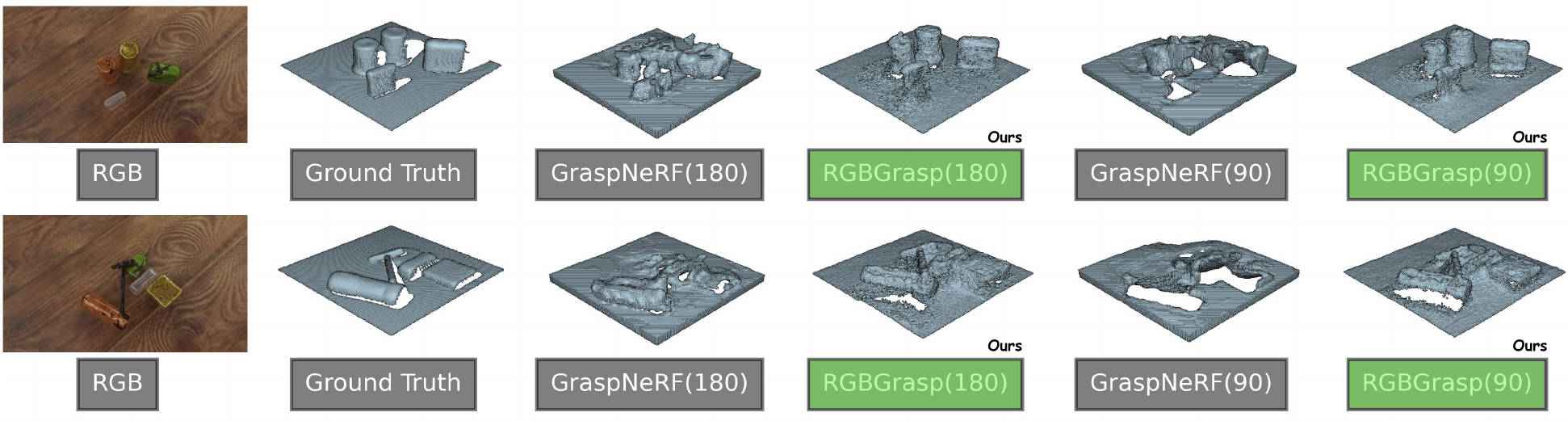}
\end{center}
\vspace{-0.5cm}
\caption{\textbf{Comparison Between GraspNeRF and RGBGrasp (Ours) in Terms of Reconstructed Point Clouds.} 
This figure presents a comparative analysis of point cloud reconstructions obtained using RGBGrasp and GraspNeRF under different trajectories, specifically, $90^\circ$ and $180^\circ$. 
The upper row provides a comparison in the "Packed" scenario, while the lower row presents the comparison in the "Pile" scenario.}
\vspace{-0.6cm}
\label{fig:experiment1}
\end{figure*}


\subsubsection{Speedup NeRF training}
In pursuit of expediting the processes of training and execution, we employ the hash encoding methodology as introduced in the work by Müller et al. \cite{muller2022instant}. Within the context of a fully connected network denoted as $m(y, \Phi)$, hash encoding pertains specifically to the encoding component, denoted as $y = enc(x, \theta)$. Rather than exclusively training the network's parameters represented by $\Phi$, we extend the training process to include the encoding weights, denoted as $\theta$. These encoding weights encompass multiple levels of multi-resolution feature maps, totaling $L$ levels. Each level operates independently and conceptually retains feature vectors situated at grid vertex locations. The final output is derived through the concatenation of feature maps that exhibit varying resolutions.

Furthermore, in contrast to the original NeRF methodology \cite{nerf2020} which employs a dual-network approach consisting of a ``coarse'' network and a ``fine'' network, our approach follows the proposal sampling technique introduced in MIP-NERF 360 \cite{barron2022mip}. Specifically, we utilize a proposal Multi-Layer Perceptron (MLP) to predict volumetric density, subsequently transformed into a proposal weight vector. These proposal weights are employed to sample intervals, which are subsequently supplied to the NeRF MLP. The NeRF MLP, in turn, predicts its own weight vector and color estimates, which are utilized in the image rendering process.

\subsection{Grasp Pose Detection}
We evaluate the geometry estimation of RGBGrasp by combining it with the downstream grasping task and analyze the grasping performance. Given the reconstructed point cloud from RGBGrasp, we can pass it to the end-to-end grasping network AnyGrasp~\cite{fang2023anygrasp} to directly predict the 6-DOF grasping poses. Then the robot arm will execute the target rotation and position of the grasping pose selected from filtered pose candidates.

\subsection{Implementation Details}
\subsubsection{Robot Arm Movement Planning}
\label{movement_planning}
Since the camera on the robot gripper continually captures images while the robot arm approaches the scene, we find that the arm's trajectory cannot be linear from the initial position to the objects. A linear trajectory causes an extremely narrow field of view of the whole scene, which will downgrade the quality of the reconstructed point cloud to a great extent. We observe up to 4cm to 5cm estimated error for the depth of the whole scene, which is unacceptable for our grasping task.

In view of that, we design an approach strategy. 
The projected trajectory to the table plane (xy plane in the world coordinate system) is about a 90-degree portion of the circle, of which the center is around the scene's center. \rebuttal{Please refer to Appendix \ref{appendix:angle_selection} for the results of the variation in grasping performance with respect to the view angle.}
To ensure the robot arm not to get contact with the objects accidentally in advance, we set a height threshold for approaching the process of about 40cm. 
For each point in the trajectory, $(x,y)$ is linearly distributed on the 90-degree portion of the circle and $z$ is linearly distributed from the initial height and the height threshold. We only captured 12 views in total along the trajectory, with each view looking at the scene. 

It should be noted that our method is \textbf{not} limited to a specific trajectory. In certain challenging scenarios, such as environments with obstacles, our method can still operate effectively within a smaller field of view than previous works. Hence, we have the flexibility to adapt the motion trajectory to different daily scenes.

\subsubsection{NeRF Training Details} We utilize the popular NerfStudio~\cite{nerfstudio} library for real-time, online training of NeRF. To enable NerfStudio's online training, we also learned from NerfBridge~\cite{yu2023nerfbridge}, which supports streaming the captured images on ROS to NeRF training. During training, in each step we sample 8192 rays across the batch, and the NeRF model is trained for 1200 iterations considering both the reconstruction quality and the time taken to estimate the depth, with a learning rate of 6e-4 and a weight decay of 1e-2. Our methods are evaluated on an NVIDIA RTX 3090 GPU.

\section{Experiments}
\label{sec:experiments}

In this section, we demonstrate the performance of our method through experiments conducted in simulation and real robot environments, respectively.


\begin{figure}[t]
\begin{center}
    \includegraphics[width=\linewidth, height=3cm]{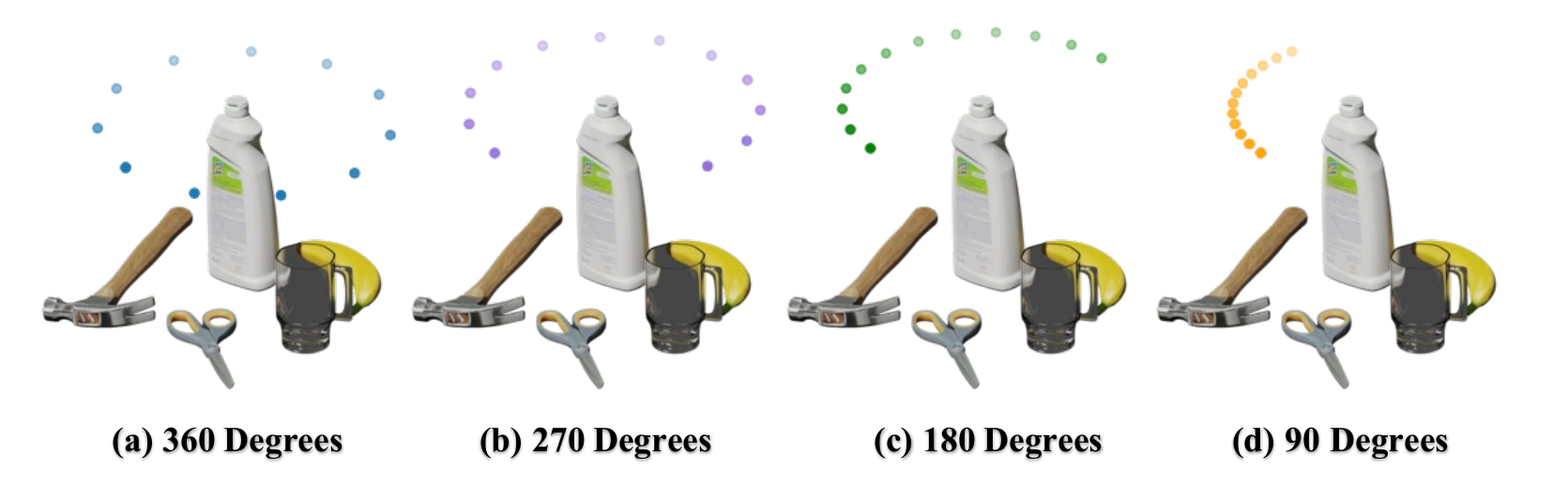}
\end{center}
\vspace{-0.4cm}
\caption{\textbf{Visualization of Trajectories with Different View Ranges}. The visualization exclusively displays the covered areas achieved through multiple views. In the original configuration (on the left (a)), views are evenly distributed to cover a full 360° at a consistent height. We successively reduce the view range to 270°, 180° and 90°, resulting in the trajectories shown in (b), (c) and (d).}
\vspace{-0.6cm}
\label{fig:different_trajectories}
\end{figure}

\begin{figure*}[t]
\begin{center}
    \includegraphics[width=0.8\linewidth, height=5.1cm]{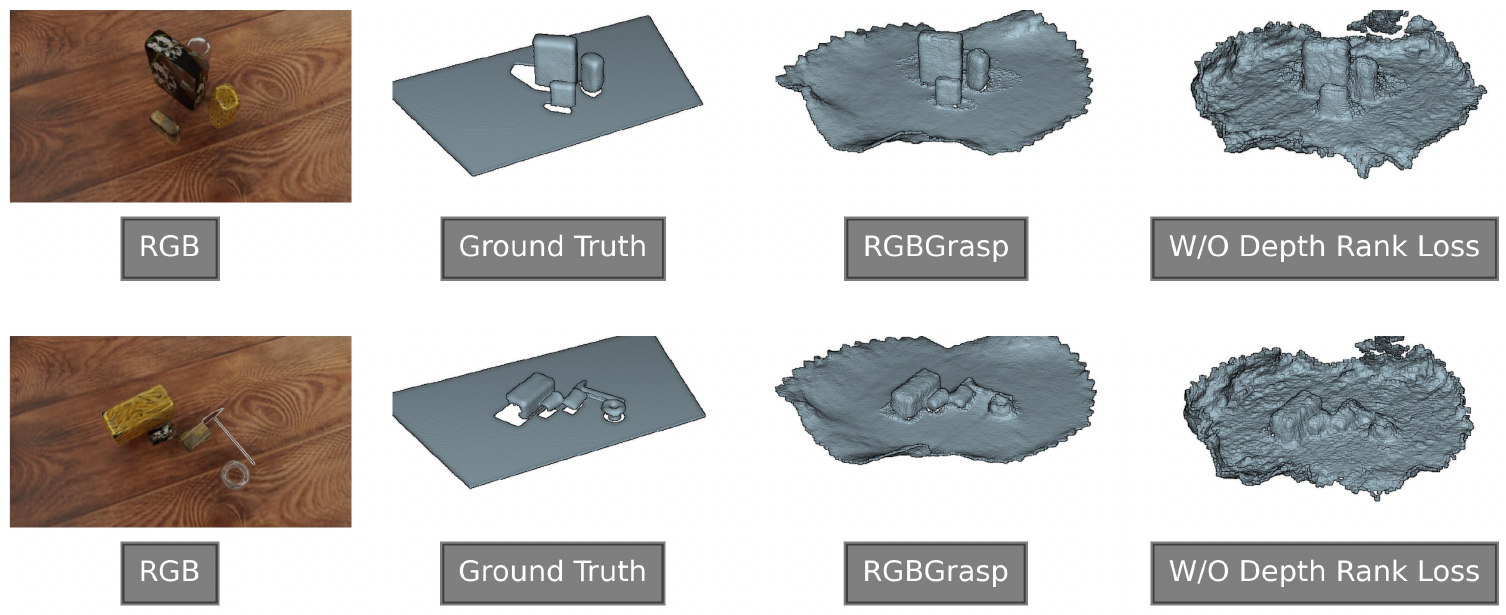}
\end{center}
\vspace{-0.4cm}
\caption{\textbf{Visualization of Ablation Studies.} This figure illustrates a comparative analysis between RGBGrasp and the ablation version of our method regarding the quality of 3D geometry reconstruction.}
\vspace{-0.6cm}
\label{fig:experiment3}
\end{figure*}

\subsection{Experiment Setup}

\subsubsection{Simulation Setup}

We train and evaluate our method in Pybullet~\cite{coumans2021}, and all the synthetic data we used to train our method in simulation is generated by Blender~\cite{blender}.
We adapted the simulation setups in VGN and GraspNeRF, which constrain all the object positions within a tabletop workspace of size $30 \times 30 \times 30 cm^{3}$. We simulate two types of scenes in this work. In the \textbf{pile} scene generation steps, we load the CAD models into the simulator and simulate their freefall from a point just above the desktop. We wait until none of the objects move in the scene and remove objects that fall out of the workspace. In the \textbf{packed} scene, we generate each object in an upright position on the table. The CAD models and poses of the objects left on the table are then recorded. We randomize the material of the objects to diffuse, transparent, or specular, then pass them to Blender along with their poses to generate photorealistic synthetic data.

\subsubsection{Real Robot Setup}

We attach a RealSense D415 RGB-D camera to the gripper's wrist of the Franka Emika Panda robot arm and leverage only the RGB observation from the camera while moving to train our method. The robot arm follows our designed trajectory and dynamically predicts grasp poses in real time. Similarly, the objects are also placed in a tabletop workspace of $30 \times 30 \times 30 cm^{3}$.

\subsubsection{Object Set}
To keep a fair comparison with the baseline methods, we use 56 hand-scaled object meshes from ~\cite{breyer2021volumetric} dataset which is never seen by the baseline method for evaluation. We simulate these objects and randomize their textures and materials as aforementioned. 

\subsection{Baseline Methods}

We compared our approach with various baseline methods that detects 6DoF grasp pose, including RGB-based methods and RGB-D-based methods.

\textbf{RGB-based methods:}
\begin{itemize}
    \item \textbf{GraspNeRF~\cite{dai2023graspnerf}}. An end-to-end grasping network that can deal with transparent and specular objects which utilizes a TSDF volume integrated from multi-view RGB-only images from fixed view points distributed on the circumference.
    \item \textbf{InstantNGP~\cite{muller2022instant} + AnyGrasp~\cite{fang2023anygrasp}}. We replace our NeRF model with InstantNGP, and export the estimated point cloud to AnyGrasp module to detect the grasp poses.
    
\end{itemize}

\textbf{RGB-D-based methods:}
\begin{itemize}
    \item \textbf{\rebuttal{Single/Fused} simulated raw point cloud~\cite{dai2022domain} + AnyGrasp~\cite{fang2023anygrasp}}. Following the activate stereo techniques of RealSense D415 RGB-D camera, we replicate the process by rendering left and right IR images of the scene using Blender. Subsequently, employing stereo matching techniques~\cite{SGM2008}, we obtain the depth information of the scene and reconstruct the single-view point cloud. \rebuttal{To ensure a fairer comparison between different methods, for RGB-D-based methods, if the same trajectory is adopted, the individual point clouds obtained along that trajectory are fused into a single point cloud for inference.}
\end{itemize}

\subsection{Evaluation Metrics}
We measure our performance with the following evaluation metrics:
\begin{itemize}
    \item \textbf{Success Rate (SR)} The number of successful grasp divided by the number of total grasp attempts
    \item \textbf{Declutter Rate (DR)} The percentage of removed objects among all objects in the scene. The reported declutter rates are averages over all rounds. 
    \item \textbf{Depth RMSE} We also measure the error of depth prediction for all methods, which could be formulated as: $RMSE = \sqrt{\frac{1}{n} \sum_{i=1}^{n} (y_i - \hat{y}_i)^2}$.
\end{itemize}

\subsection{Simulation Grasping Experiments}

We conduct two sets of experiments in simulation environment to illustrate the stability and effectiveness of our method. The first set of experiments shown in Sec.~\ref{sec:exp_type1} reveals the stability of our method when the view range decreases, while the second set of experiments in Sec.~\ref{sec:exp_type2} demonstrates the effectiveness compared to other baselines.

In all scenarios, we carry out 200 rounds of decluttering experiments, encompassing both pile and packed scenes, and involving a diverse range of materials such as diffuse, transparent, and specular objects.

During each trial in a single round of experimentation, the robot arm performs a grasping action followed by the removal of one object. This process is repeated until one of the following conditions is met: (1) the entire workspace is cleared of objects, (2) the grasp detection system fails to detect a successful grasp, or (3) two consecutive grasping failures occur.

\subsubsection{Clutter Removal Experiments under Trajectories with Various View Ranges}
\label{sec:exp_type1}
In this study, we evaluate the performance of RGBGrasp and GraspNeRF in the context of grasping tasks using various camera trajectories within our experimental framework. The trajectories are visualized in Fig.~\ref{fig:different_trajectories}. Our findings, reported in Tab.~\ref{ClutterSim_FlyingGripper}, indicate that RGBGrasp generally outperforms GraspNeRF in all trajectories.
As the trajectory angle decreases from 360° to 90°, RGBGrasp maintains a high success rate, while degradation can be witnessed for GraspNeRF. Additionally, even in scenarios with only a 90° field of view, RGBGrasp consistently delivers competitive performance.

In this experiment, we also assess the perceptual capabilities of our proposed 3D reconstruction system in the context of grasping scenarios. Given that grasping tasks heavily rely on accurate depth perception, we conduct a comparative evaluation of our algorithm's depth reconstruction performance against that of GraspNeRF.

The results of these comparisons are summarized in Tab.~\ref{ClutterSim_FlyingGripper_Perception}. It is noteworthy that, in the majority of cases, our algorithm exhibits superior performance compared to GraspNeRF. An interesting observation is that GraspNeRF employs a TSDF (Truncated Signed Distance Function) as an intermediate representation for the 3D environment, resulting in a substantially higher number of generated point clouds compared to our approach. This difference in representation leads to a discrepancy between the visualizations and the quantitative results presented. Even though our algorithm demonstrates markedly improved qualitative performance in Fig.~\ref{fig:experiment1}, the quantitative results do not reflect a substantial difference.

Qualitative results presented in Fig.~\ref{fig:experiment1} align with the quantitative findings reported in Tab.~\ref{ClutterSim_FlyingGripper}.

\begin{table}[!htb]
    \vspace{-0.3cm}
    \centering
    \caption{Results of Clutter Removal Experiments in Simulation \textbf{with Various View Ranges}}
    \label{ClutterSim_FlyingGripper}
\resizebox{1\linewidth}{!} 
{\begin{tabular}{ c|c c c c c } 
        \hline
         & \multicolumn{2}{c}{Pile} & \multicolumn{2}{c}{Packed} \\ 
        \hline
         & SR (\%)  & DR (\%)  & SR (\%) & DR (\%)  \\ 
         \hline
         GraspNeRF (90\si{\degree})  & 17.4 & 9.4 & 28.1 & 23.6 \\
         \hline
         GraspNeRF (180\si{\degree}) & 37.6  & 25.1 & 69.1 & 69.6 \\
        \hline
        GraspNeRF (270\si{\degree}) & 60.3  & 38.9 & 80.6 & 77.7 \\
        \hline
         GraspNeRF (360\si{\degree}) & 70.0 & 45.1 & 81.5 & 80.6 \\
        \hline
        \hline
        RGBGrasp (90\si{\degree}) &  \textbf{86.7} & \textbf{81.5} & \textbf{84.8} & \textbf{86.0} \\
        \hline
        RGBGrasp (180\si{\degree}) &  \textbf{84.5} & \textbf{77.0} & \textbf{86.2}  & \textbf{88.0} \\
        \hline
        RGBGrasp (270\si{\degree}) &  \textbf{84.6} & \textbf{77.5} & \textbf{85.8}  & \textbf{86.5} \\
        \hline
        RGBGrasp (360\si{\degree}) &  \textbf{85.9} & \textbf{80.0} & \textbf{85.3} & \textbf{86.5} \\
        \hline
    \end{tabular}}
    \vspace{-0.25cm}
\end{table}

\begin{table}
    \vspace{-0.1cm}
    \centering
    \caption{Depth Error of Reconstruction Results of Clutter Removal Experiments in Simulation
            with \textbf{Various View Ranges}}
    \label{ClutterSim_FlyingGripper_Perception}
\resizebox{1\linewidth}{!} 
{
    \begin{tabular}{ c|c c c c c c} 
        \hline
        &\multicolumn{1}{c}{Pile} & 
        \multicolumn{1}{c}{Size} &
        \multicolumn{1}{c}{Packed} & \multicolumn{1}{c}{Size} \\ 
        \hline
         GraspNeRF (90\si{\degree}) & \bm{$9.61\times 10^{-3}$} & $261935$ & $1.00\times 10^{-2}$ & $373179$ \\
         GraspNeRF (180\si{\degree}) & $9.05\times 10^{-3}$ & $276607$ & $9.93\times 10^{-3}$ & $379318$\\
         GraspNeRF (270\si{\degree}) & $8.40\times 10^{-3}$ & $278363$ & $8.67\times 10^{-3}$ & $346129$\\
         
         GraspNeRF (360\si{\degree}) &  $8.07\times 10^{-3}$ & $272098$ & $8.42\times 10^{-3}$ & $343121$\\
        \hline
        RGBGrasp (90\si{\degree}) & $9.94 \times 10^{-3}$ & $62806$ & \bm{$9.09\times 10^{-3}$} & $66967$\\
        RGBGrasp (180\si{\degree}) & \bm{$6.58 \times 10^{-3}$} & $62893$ & \bm{$7.20\times 10^{-3}$} & $66973$\\
        RGBGrasp (270\si{\degree}) & \bm{$6.89\times{10^{-3}}$} & $62396$ & \bm{$6.91\times{10^{-3}}$} & $65922$ \\
        
        RGBGrasp (360\si{\degree}) & \bm{$7.12\times 10^{-3}$} & $62903$ & \bm{$7.58\times 10^{-3}$} & $66974$\\
        \hline
     
    \end{tabular}}
    \vspace{-0.5cm}
\end{table}
\subsubsection{Clutter Removal Experiments with Approaching Trajectory}
\label{sec:exp_type2}

In this experimental setup, we conducted simulations following a predefined trajectory. Our evaluation included a comparison of RGBGrasp against several baseline methods, namely Instant-NGP with AnyGrasp as the grasp module (Instant-NGP+AnyGrasp), \rebuttal{single simulated point cloud with AnyGrasp (Single Sim Point Cloud+AnyGrasp)}, \rebuttal{fused simulated point cloud with AnyGrasp (Fused Sim Point Cloud+AnyGrasp)} and our method (RGBGrasp). GraspNeRF can only work from fixed viewpoints, and it is unable to reconstruct a reasonable TSDF volume when performing perception directly along the trajectory. Therefore, we do not compare our method with GraspNeRF in this regard.

The trajectory, as detailed in the implementation section, provided a relatively narrow 90-degree field of view, which presents a limitation for NeRF-based approaches in reconstructing fine-grained object geometry. Due to this constrained view and the trajectory's randomized and flexible nature, Instant-NGP fails to effectively reconstruct the geometry of the scene. The point cloud is filled with artifacts such as floating points, which prevents grasp modules like AnyGrasp from providing accurate grasp poses.

We evaluate the performance in two settings: the first number in each cell represents the metric in diffuse scenes (containing only diffuse objects), and the second in mixed scenes (containing diffuse, specular and transparent objects), as presented in Tab.~\ref{tab.exp2_diffuse_and_mixed}. In scenes consisted of diffuse objects, the performance of RGBGrasp is comparable to RGB-D based method. For mixed scenes, accuracy of simulated depth downgrades greatly because of reflection and refraction of light, which leads to a significant decrease in SR and DR. \rebuttal{Although multi-view RGB-D fusion can improve SR and DR to certain extent in mixed scenes, the inability to accurately perceive point clouds of transparent objects still limits the performance compared to our RGB-based method.} In contrast, our method is practically unaffected in these scenes.


\begin{table}[!htb]
    \vspace{-0.3cm}
    \centering
    \caption{Results of Clutter Removal Experiments in Simulation \textbf{with Viewpoints on Approaching Trajectory} \\ Each Cell: Diffuse/Mixed}
\resizebox{1\linewidth}{!}
{\begin{tabular}{c|c|cccc}
\hline
  &          & \multicolumn{2}{c}{Pile} & \multicolumn{2}{c}{Packed} \\ \hline
  & modality & SR (\%)  & DR (\%)       & SR (\%)  & DR (\%)         \\ \hline

\begin{tabular}[c]{@{}c@{}}Single Sim Point Cloud \\ + AnyGrasp\end{tabular} & RGB-D  &83.9/76.2&75.5/65.0&\textbf{81.9}/66.1&81.5/55.0     \\ \hline 

\begin{tabular}[c]{@{}c@{}}\rebuttal{Fused Sim Point Cloud} \\ \rebuttal{+ AnyGrasp}\end{tabular} & \rebuttal{RGB-D}&\rebuttal{\textbf{84.0}/78.4} & \rebuttal{74.5/62.5} & \rebuttal{81.6/77.7} &\rebuttal{81.5/77.0}\\ \hline \hline

RGBGrasp& RGB& 83.7/\textbf{82.2}       &     \textbf{81.0}/\textbf{80.0}&81.5/\textbf{80.9}&\textbf{82.5}/\textbf{82.0}\\ \hline
\end{tabular}}
    \label{tab.exp2_diffuse_and_mixed}
    \vspace{-0.7cm}
\end{table}


\subsection{Real Grasping Experiments}
\label{sec:real_grasp}
To further evaluate the grasping performance in real-world scenarios, we conduct 15 rounds of clutter removal experiments on both pile and packed scenes, following the protocol of simulation experiments. Each scene consists of a combination of five randomly selected daily objects, including objects with diffuse, transparent, and specular materials. The round ends with failure if two consequent grasps fails, or ends with success with all objects cleared. 

We compare our RGBGrasp with \rebuttal{three} baselines, namely GraspNeRF, AnyGrasp with a monocular topdown RGB-D image \rebuttal{and AnyGrasp with fused RGB-D images}. Results are shown in Tab.~\ref{tab.realworld}. Given that each scene contains at least one transparent daily object, AnyGrasp's performance downgrades much because of the raw, corrupted depth image of RealSense D415. However, RGBGrasp reconstructs more accurate and fine-grained point cloud than RGB-D cameras, which boosts the final success rate of the grasping tasks. GraspNeRF shows lack of generalization since the daily objects we randomly picked are diverse and some of them are out of distribution of VGN dataset. Furthermore, the training process of GraspNeRF only takes into account grasping using a flying gripper in the simulator, thus resulting in some generated grasp poses that are not practical for execution on real robots. For example, the predicted side grasps in the packed scenes can easily cause the robotic arm to reach its joint limits when the object is either slightly close or slightly far from the robot base.

\rebuttal{For clutter removal experiments in more complex and real-world scenarios, you can refer to Appendix \ref{appendix:complex} to see the experiment details. The videos are on our webpage: \url{https://sites.google.com/view/rgbgrasp}.}
\begin{table}[!htb]
\vspace{-0.3cm}
\centering
\caption{Results of Clutter Removal Experiments in Real World}
\resizebox{1\linewidth}{!}
{\begin{tabular}{c|c|cccc}
\hline
&  & \multicolumn{2}{c}{Pile} & \multicolumn{2}{c}{Packed} \\ \hline
            & modality & SR (\%)         & DR (\%)        & SR (\%)          & DR (\%)          \\ \hline
GraspNeRF   & RGB      &   70.1 (54/77)         &   46.7 (7/15)         &     66.7 (52/78)         &  40.0 (6/15)          \\ \hline
\makecell{RealSense D415 \\ + AnyGrasp \rebuttal{(Single Point Cloud)}} & RGB-D &     61.4 (43/70)        &   33.3 (5/15)         &     59.7 (43/72)        &    33.3 (5/15)        \\ \hline 
\makecell{\rebuttal{RealSense D415} \\ + \rebuttal{AnyGrasp (Fused Point Cloud)}} & \rebuttal{RGB-D} &     \rebuttal{71.4 (65/91)}        &   \rebuttal{73.3 (11/15)}         &     \rebuttal{66.7 (56/84)}        &    \rebuttal{46.7 (7/15)}        \\ \hline \hline
RGBGrasp    & RGB      &      \textbf{84.1 (74/88)}       &    \textbf{93.3 (14/15)}        &      \textbf{77.1 (64/83)}        &     \textbf{66.7 (10/15)}        \\ \hline 
\end{tabular}}
\label{tab.realworld}
\vspace{-0.6cm}
\end{table}

\subsection{Ablation Studies}

For ablation analysis, we conduct clutter removal experiments on mixed scenes in simulation between RGBGrasp and RGBGrasp without depth rank loss. 

The corresponding qualitative results are illustrated in Fig.~\ref{fig:experiment3}, where we observe that even under these challenging conditions, our method managed to produce reasonable 3D representations of the scenes. As we can see, there are more artifacts in the reconstructed point cloud without depth supervision, especially on the boundary of the point cloud. However, most grasp modules rely on local geometry features to generate grasp poses, and these artifacts would cheat grasp modules to detect grasps around them. This phenomenon is more likely to occur in pile scenes, as objects in pile scenes have smaller dimensions and are more susceptible to artifacts. The quantitive results are shown in Tab.~\ref{tab.ablations}.

\begin{table}[!htb]
\vspace{-0.3cm}
\centering
\caption{Ablation Studies on Clutter Removal Experiments \textbf{with Viewpoints on Approaching Trajectory} in Simulation}
\begin{tabular}{c|cccc}
\hline

                                                                        & \multicolumn{2}{c}{Pile} & \multicolumn{2}{c}{Packed} \\ \hline
                                                                        & SR (\%)     & DR (\%)    & SR (\%)         & DR (\%)  \\ \hline
\begin{tabular}[c]{@{}c@{}}RGBGrasp \\ w/o depth rank loss\end{tabular} &      79.3       &    75.5       &       79.7       &      81.0       \\ \hline \hline
RGBGrasp                                                                &    \textbf{82.2}         &   \textbf{80.0}         &  \textbf{80.9}            &   \textbf{82.0}          \\ \hline
\end{tabular}
\label{tab.ablations}
\vspace{-0.55cm}
\end{table}
 
\section{Conclusion \& Future Works}
\label{sec:conclusion}

Our algorithm pioneers accurate 3D geometry reconstruction for typical grasping trajectories by incorporating prior knowledge from a monocular depth model. To achieve real-time reconstruction, we combined hash encoding and proposal sampling. Our experiments demonstrate superior performance in various trajectories for perception and grasping tasks. However, challenges remain in certain scenarios, prompting us to explore integrating prior knowledge into object detection for enhanced accuracy in future research.





\bibliographystyle{IEEEtran}
\bibliography{IEEEabrv,reference}
\vspace{-0.5cm}
\appendices

\section{The variation in grasp performance with respect to the view angle}
\label{appendix:angle_selection}

As mentioned in Sec. \ref{movement_planning}, we chose a trajectory with 90-degree view angle. The view angle of the trajectory is an important factor that affects the grasping performance. 

We conducted tests using the experiment protocol in Sec. \ref{sec:experiments} and got results below:

\begin{figure}[htbp]
	\centering
	\begin{minipage}{\linewidth}
		\centering
        
		\includegraphics[width=\linewidth]{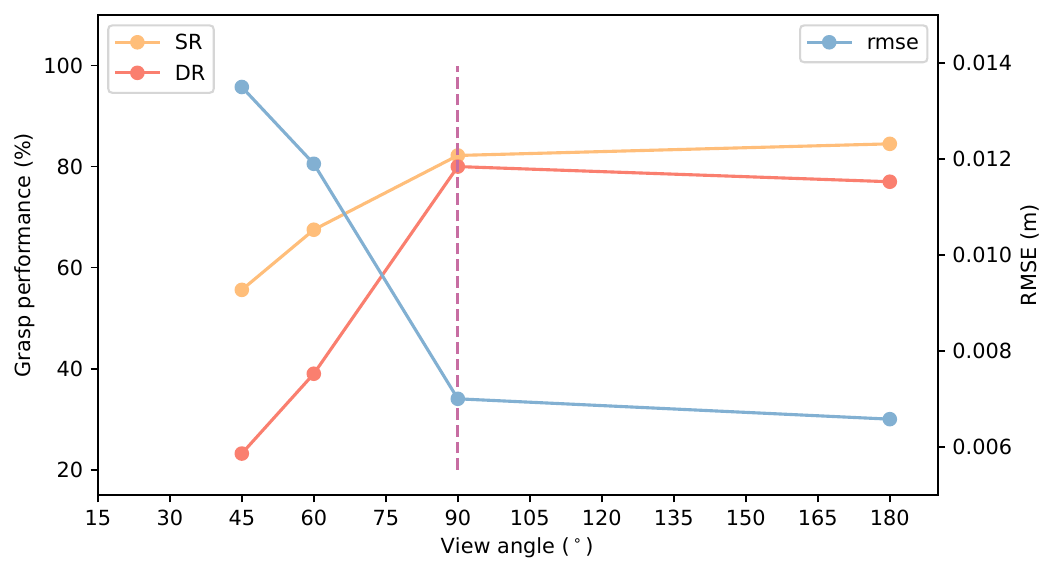}
        \vspace{-0.8cm}
		\caption{SR (success rate), DR (declutter rate) and reconstruction RMSE of RGBGrasp with approaching trajectories of different view angles in \textbf{pile} scenes.}
		\label{fig:pile_trajectory}
	\end{minipage}
	
	\begin{minipage}{\linewidth}
		\centering
		\includegraphics[width=\linewidth]{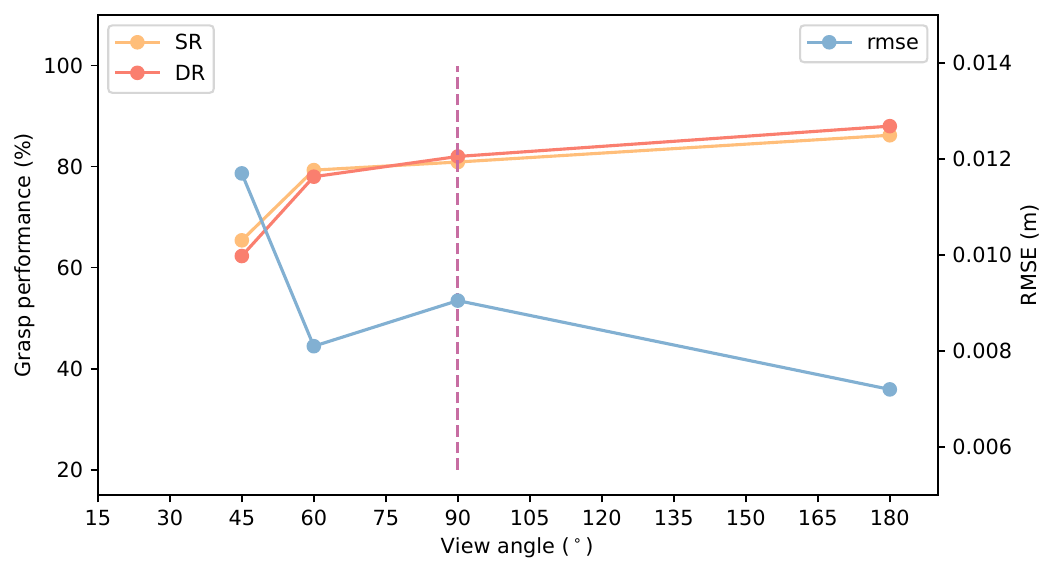}
        \vspace{-0.8cm}
		\caption{SR (success rate), DR (declutter rate) and reconstruction RMSE of RGBGrasp with approaching trajectories of different view angles in \textbf{packed} scenes.}
		\label{fig:pack_trajectory}
	\end{minipage}
\end{figure}

According to Fig. \ref{fig:pile_trajectory} and \ref{fig:pack_trajectory}, we concluded that $90^\circ$ is a proper trajectory setting for our method based on two points:

\begin{itemize}
    \item There is not much performance gain when the view angle exceeds $90^\circ$ in both pile and packed scenes.
    \item $90^\circ$ is a relatively narrow angle compared to $180^\circ$, which makes our method more flexible as it does not need to occupy so much space.
\end{itemize}

\vspace{-0.3cm}

\section{Real world experiments in complex scenarios}
\label{appendix:complex}

\begin{figure}[htbp]
\begin{center}
    \includegraphics[width=\linewidth]{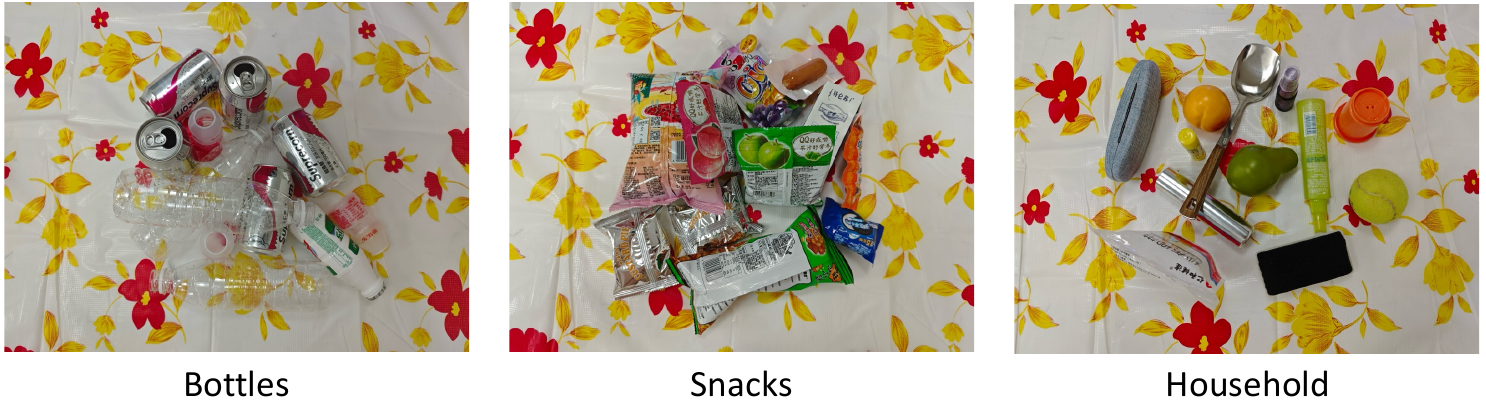}
\end{center}
\vspace{-0.4cm}
\caption{Object types of complex scenarios}
\vspace{-0.4cm}

\label{fig:types}
\end{figure}

To better demonstrate the superiority of our method in extreme scenarios, we conducted further real-world experiments on several baselines described in Sec. \ref{sec:real_grasp}. In these experiments, the objects are densely stacked, resulting in a high level of clutter. The types of objects are shown in Fig. \ref{fig:types}.

Due to the large number of objects exceeding the working range of GraspNeRF ($30\times30\times30\ cm^3$), it is challenging for GraspNeRF to achieve satisfactory performance. Therefore, we only conducted experiments between AnyGrasp+Single Point Cloud (namely RGB-D Single), AnyGrasp+Fused Point Cloud (namely RGB-D fused) and our method. The \textbf{attempt-centric} success rate (defined as the ratio of the number of successful grasp attempts to the total number of grasp attempts) and \textbf{object-centric} success rate (defined as the ratio of the number of successfully grasped objects to the total number of objects) are as follows:

\vspace{-0.3cm}

 \begin{figure}[htbp]
\begin{center}
    \includegraphics[width=0.8\linewidth]{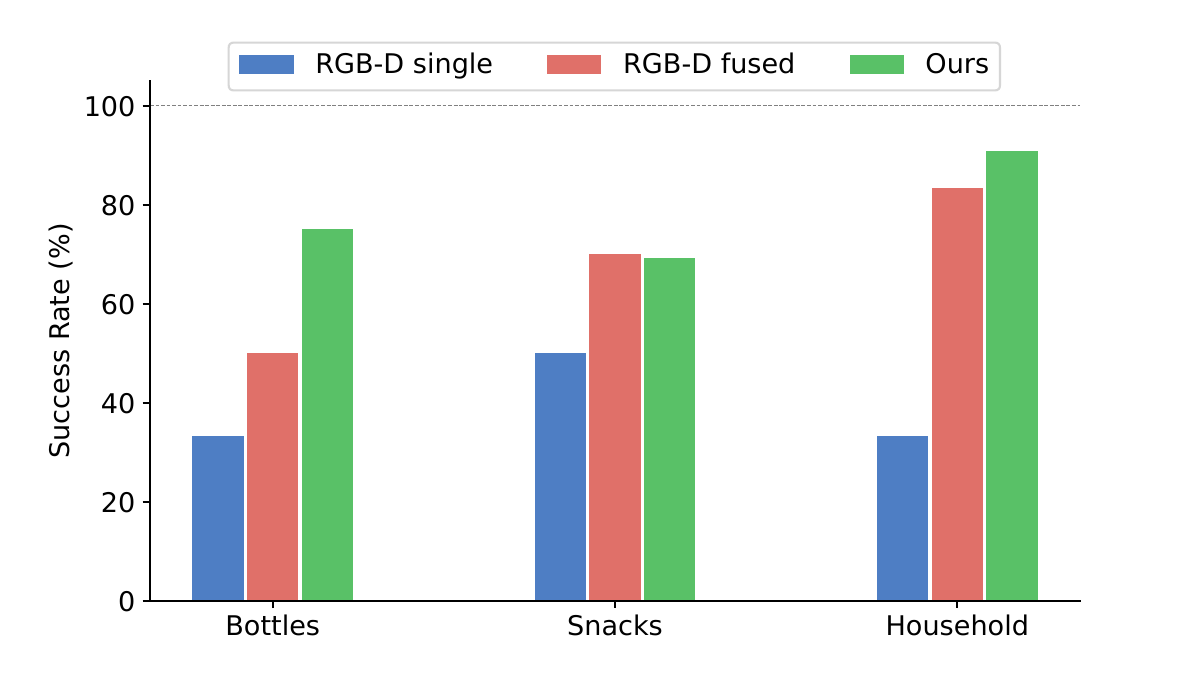}
\end{center}
\vspace{-0.8cm}
\caption{\textbf{Attempt-Centric Success Rate of Complex Scenarios.}}
\label{fig:attempt_centric}
\vspace{-0.6cm}
\end{figure}

 \begin{figure}[htbp]
\begin{center}
    \includegraphics[width=0.8\linewidth]{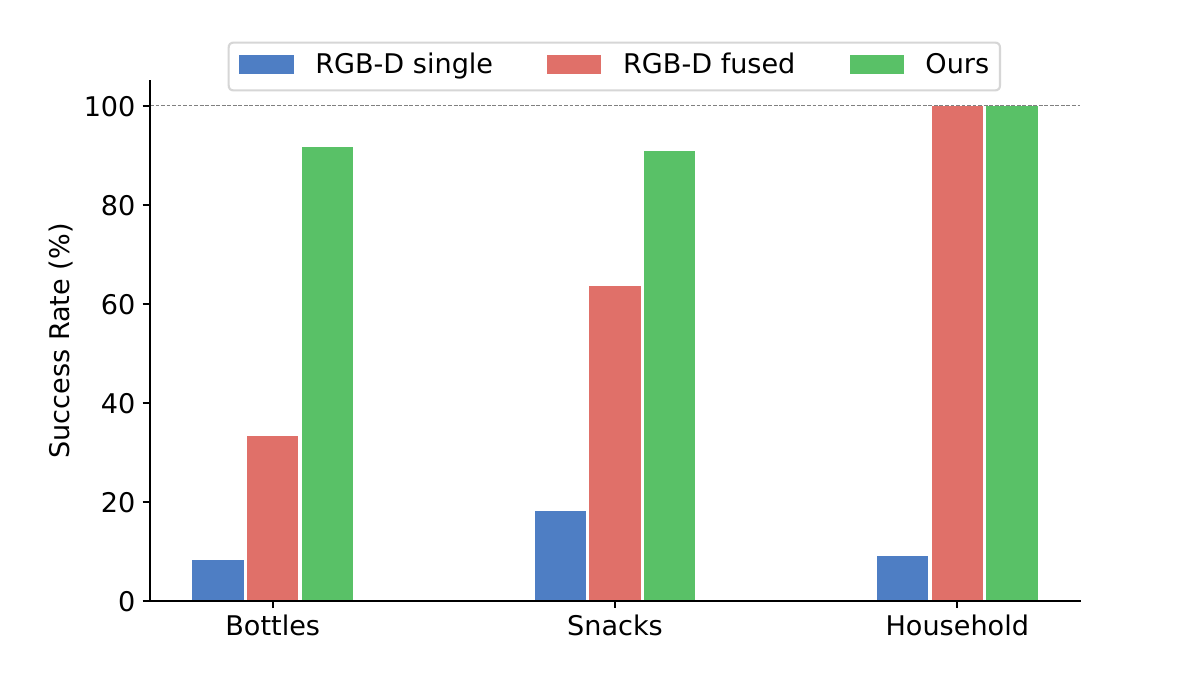}
\end{center}
\vspace{-0.8cm}
\caption{\textbf{Object-Centric Success Rate of Complex Scenarios.}}
\label{fig:object_centric}
\end{figure}
\vspace{-0.3cm}
The grasp performance of our method exceeds RGB-D method when considering reflections and transparency (see "Bottles") and is comparable to multi-view RGB-D method in other normal scenes (see "Snacks" and "Household").


\end{document}